\documentclass[letterpaper]{article} 
\usepackage{aaai2026}  
\usepackage{times}  
\usepackage{helvet}  
\usepackage{courier}  
\usepackage[hyphens]{url}  
\usepackage{graphicx} 
\usepackage{natbib}  
\usepackage{caption} 
\usepackage{algorithm}

\usepackage{amsmath}
\usepackage{mdframed}
\usepackage{booktabs}
\usepackage{multirow}
\usepackage[noend]{algpseudocode}

\usepackage{newfloat}
\usepackage{listings}
\DeclareCaptionStyle{ruled}{labelfont=normalfont,labelsep=colon,strut=off} 
\floatstyle{ruled}
\newfloat{listing}{tb}{lst}{}
\floatname{listing}{Listing}
\title{MRGSEM-Sum: An Unsupervised Multi-document Summarization Framework based on Multi-Relational Graphs and Structural Entropy Minimization}
\author{
    Yongbing Zhang, Fang Nan, Shengxiang Gao, Yuxin Huang, Kaiwen Tan*, Zhengtao Yu*
}
\affiliations{
    Faculty of Information Engineering and Automation, Kunming University of Science and Technology, Kunming, China \\
    Yunnan Key Laboratory of Artificial Intelligence, Kunming, China

    * Corresponding authors: kwtan@kust.edu.cn, ztyu@hotmail.com
}

\usepackage{bibentry}

\begin{document}

\maketitle

\begin{abstract}
The core challenge faced by multi-document summarization is the complexity of relationships among documents and the presence of information redundancy. Graph clustering is an effective paradigm for addressing this issue, as it models the complex relationships among documents using graph structures and reduces information redundancy through clustering, achieving significant research progress. However, existing methods often only consider single-relational graphs and require a predefined number of clusters, which hinders their ability to fully represent rich relational information and adaptively partition sentence groups to reduce redundancy. To overcome these limitations, we propose MRGSEM-Sum, an unsupervised multi-document summarization framework based on multi-relational graphs and structural entropy minimization. Specifically, we construct a multi-relational graph that integrates semantic and discourse relations between sentences, comprehensively modeling the intricate and dynamic connections among sentences across documents. We then apply a two-dimensional structural entropy minimization algorithm for clustering, automatically determining the optimal number of clusters and effectively organizing sentences into coherent groups. Finally, we introduce a position-aware compression mechanism to distill each cluster, generating concise and informative summaries. Extensive experiments on four benchmark datasets (Multi-News, DUC-2004, PubMed, and WikiSum) demonstrate that our approach consistently outperforms previous unsupervised methods and, in several cases, achieves performance comparable to supervised models and large language models. Human evaluation demonstrates that the summaries generated by MRGSEM-Sum exhibit high consistency and coverage, approaching human-level quality.

\end{abstract}


\section{Introduction}

Multi-Document Summarization (MDS) is designed to create compact and informative summaries from collections of documents that revolve around specific subjects, facilitating users in efficiently gaining essential information. \cite{ma2022multi, roy2023review, li2023compressed}. However, due to the large number of documents, the relationships between documents are highly complex, and there is also the issue of information redundancy among documents, leading to suboptimal results with existing methods.

\begin{figure}[h]
  \centering
  \includegraphics[width=\linewidth]{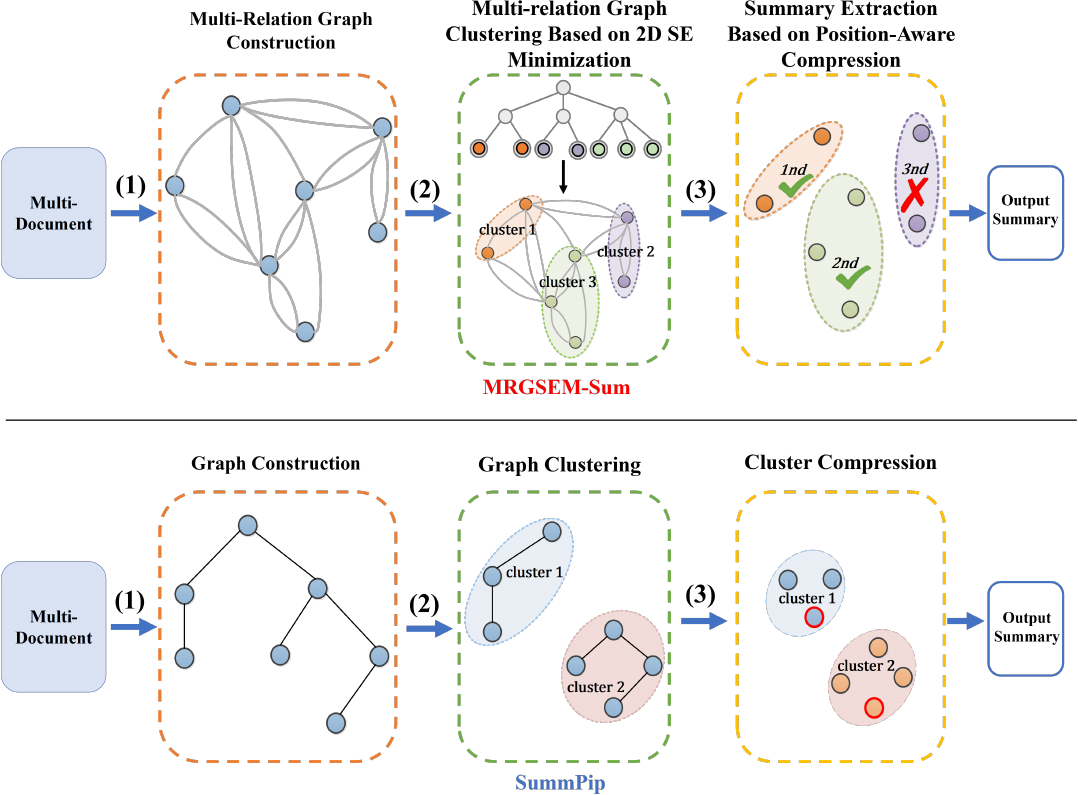}
  \caption{Model architecture comparison between MRGSEM-Sum and the existing graph-based multi-document summarization method (i.e., Sumpip)}
  \label{fig:Fig2}
\end{figure}

In response, researchers have conducted in-depth studies. Among them, unsupervised methods eliminate the need for annotated data and instead leverage the intrinsic structure and semantic relationships within document clusters to automatically generate summaries. This not only reduces data acquisition costs but also provides stronger generalization capabilities, enabling these methods to flexibly adapt to various topics and domains \cite{vogler2022unsupervised}. Therefore, unsupervised MDS methods have become a focal point in recent MDS research. Existing unsupervised methods can be categorized into autoencoders \cite{vogler2022unsupervised}, ranking-based approaches \cite{lamsiyah2021unsupervised, alambo2020topic}, and graph clustering methods \cite{zhao2020summpip, alami2021unsupervised}. While autoencoders can effectively retain detailed document information, they often struggle to filter out the core content required for summarization and face challenges with long input sequences \cite{zhao2020summpip}. Ranking-based methods, while able to identify salient sentences, are usually grounded in rigid prior assumptions, such as semantic relevance and diversity, which may not always hold in practice. In contrast, graph clustering methods, such as Sumpip \cite{zhao2020summpip}, model the complex relationships between documents by constructing a graph of relationships between sentences and then reducing redundancy among documents through clustering. This method can straightforwardly address the complexities of relationships and information redundancy in MDS and has become a current research hotspot. However, existing methods face two limitations: \textbf{First, existing graph structures are typically limited to modeling single-relation connections, constraining their ability to capture the complex and heterogeneous relationships that naturally arise in multi-document scenarios, such as semantic, discourse, and pragmatic links. Second, traditional clustering frameworks often require the number of clusters to be predefined, increasing model complexity and inflexibility.}

To address these limitations, we propose an unsupervised multi-document summarization framework based on multi-relational graphs and structural entropy minimization, named MRGSEM-Sum. Specifically, we first construct a multi-relational graph that captures semantic and discourse relationships among sentences. We then utilize a two-dimensional (2D) structural entropy (SE) minimization algorithm to cluster the sentence graph in an unsupervised manner, automatically determining the optimal number of clusters and grouping highly relevant sentences together. Finally, we introduce a position-aware compression strategy to further refine and condense each cluster, resulting in coherent and comprehensive summaries. A visual comparison of the model architectures of MRGSEM-Sum and the existing graph-based multi-document summarization method (i.e., Sumpip) is illustrated in Figure~\ref{fig:Fig2}. In summary, our key contributions are as follows:
\begin{itemize}
  \item We propose a novel unsupervised multi-document summarization framework that addresses the complexities of relationships and information redundancy in MDS.
  \item We designed a multi-relational graph construction method, combined with the 2D SE minimization clustering approach and position-aware compression mechanism, to capture the intrinsic complex relationships between documents, eliminate information redundancy, and enhance the quality of summaries.
  \item We conduct extensive experiments on four benchmark MDS datasets, including Multi-News, DUC-2004, PubMed, and WikiSum. The results demonstrate that MRGSEM-Sum outperforms state-of-the-art unsupervised baselines significantly in both automatic metrics and human evaluation, showing competitiveness with supervised methods and large language models.
\end{itemize}

\section{Methodology}\label{sec:method}
As shown in Figure \ref{fig:Fig3}, MRGSEM-Sum consists of three key modules: A. multi-relational graph construction; B. multi-relation graph clustering based on 2D SE minimization; C. summary extraction based on position-aware compression. The details are as follows:



    
\begin{figure*}[h!]
  \centering
  \includegraphics[width=0.9\textwidth]{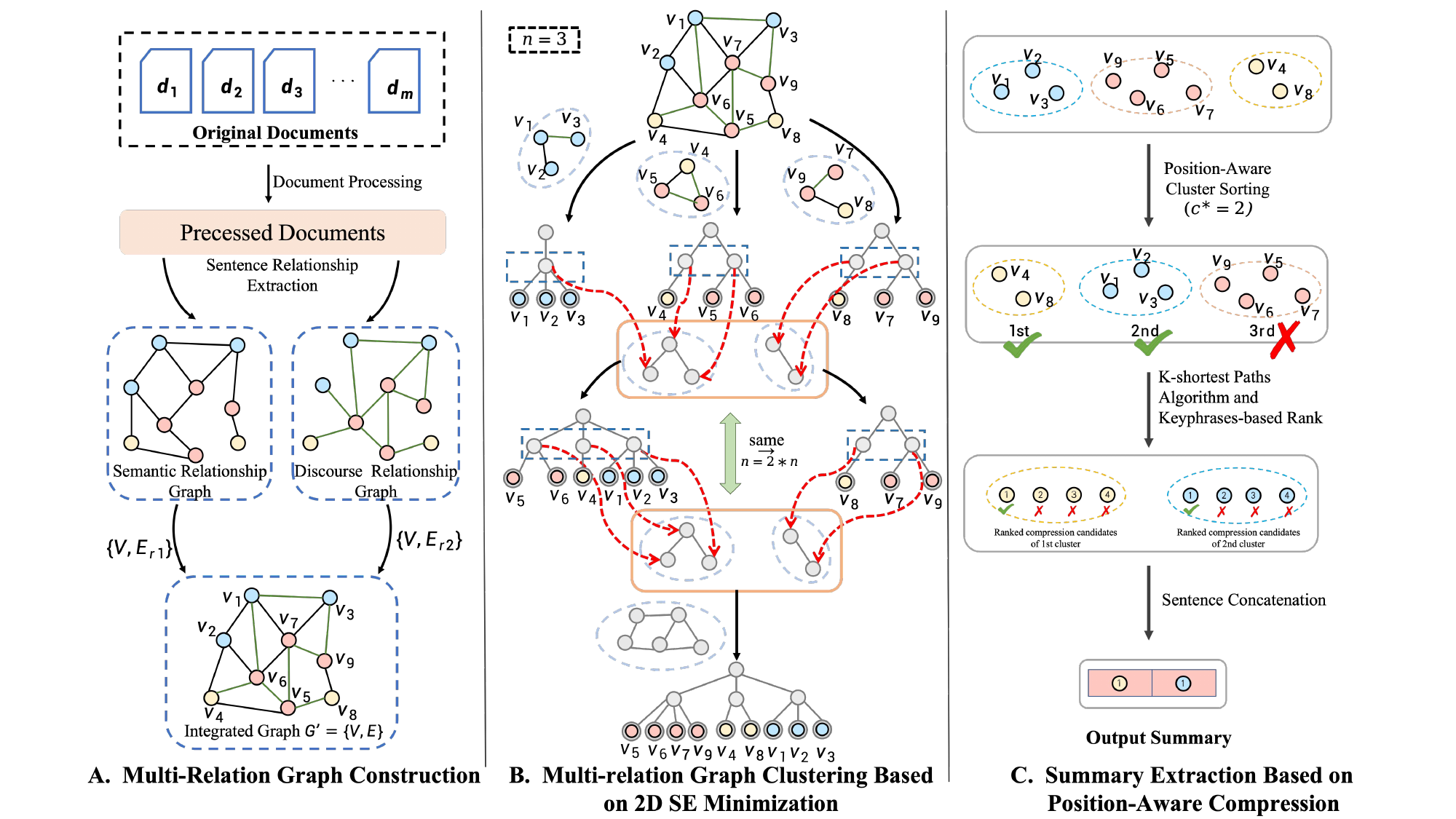}
  \caption{The framework of MRGSEM-Sum. A. Multi-relational Graph Construction. First, a multi-relational graph is constructed with sentences from multiple documents as nodes, based on sentence semantic similarity and discourse relations across the documents. B. Multi-relation Graph Clustering Based on 2D SE Minimization. Based on the constructed multi-relational graph and following the principle of SE minimization, an optimal 2D encoding tree is established. Sentences are assigned to different subtrees on the encoding tree, thereby partitioning the original multi-relational graph into subgraphs. C. Summary Extraction Based on Position-Aware Compression. The subgraphs of sentences are then compressed using a position-aware compression method to generate the final summary text.}
  \label{fig:Fig3}
\end{figure*}

\subsection{A. Multi-relational Graph Construction}\label{sec:graphcon}
To capture the complex and diverse relationships between documents in the input document cluster \( D = \{d_1, d_2, ..., d_n\} \), we first construct a multi-relational graph \( G = \{{V}, {E}_{r_1}, {E}_{r_2}\} \) utilizing both semantic relationship and discourse relationship, as illustrated in Figure \ref{fig:Fig3}A. The two types of relationships, approached from different perspectives, complement each other, thereby modeling the complex relationships between documents.

Specifically, semantic relationship based edges \( E_{r1} \) are determined by calculating the cosine similarity between frequency-inverse document frequency (TF-IDF) representations of sentences. The weight of these edges reflects the similarity at the content level, providing a quantitative measure of the potential associations between sentences. Discourse relationship based edges \( E_{r2} \) are established based on the method proposed by Cristensen et al.\cite{christensen2013towards}. We consider features such as discourse markers, coreference, and entity linking to establish discourse relationships between sentences. These edges capture not only the structural connections of sentences within the documents but also reveal their logical and rhetorical relationships. The nodes \( V \) are sentences from the input document cluster, we select a pre-trained sentence embedding model SBERT to initialize the representation of nodes, thereby obtaining high-dimensional node representations \( X\).




\subsection{Multi-relation Graph Clustering Based on 2D SE Minimization}

Existing unsupervised summarization methods often utilize clustering models on document relation graphs to aggregate similar sentences and reduce redundancy \cite{alami2021unsupervised, wang2021unsupervised}. However, these approaches typically require predefined cluster numbers, which is impractical for real-world applications. To address this limitation, we propose an adaptive clustering method based on 2D SE minimization that automatically determines the optimal cluster structure, as illustrated in Figure \ref{fig:Fig3}B.

Since the 2D SE cannot be directly applied to multi-relational graphs, we need to first merge the relationships within the multi-relational graphs to form an integrated graph. The merging process is as follows:
\begin{equation}
    A'[i][j]=\begin{cases} 
    \max(A_{E_{r_1}}[i][j], A_{E_{r_2}}[i][j]) & \text{if }  A_{E_{r_m}}[i][j] \neq 0 \\
    0 & \text{otherwise}
    \end{cases}
\end{equation}
where \(A^{'}\) represents the adjacency matrix of the integrated graph \(G^{'}=\{V, E\}\), and \(A^{'}[i][j]\) indicates the weight of the edge from node \(i\) to node \(j\). The elements \(A_{E_{r1}}[i][j]\) and \(A_{E_{r2}}[i][j]\) correspond to the weights of the edges between node \(i\) and \(j\) under the relationship types \(E_{r1}\) and \(E_{r2}\), respectively.

\begin{algorithm}[h!]
\caption{Multi-relation Graph Clustering via 2D SE minimization}
\begin{algorithmic}[1]
\Statex \textbf{Input:} Integrated graph ${G}' = ({V}, {E})$, $n$: number of nodes in each subgraph.
\Statex \textbf{Output:} A partition ${P}$ of ${V}$.
\State ${P} \gets \{ \{s\} \mid s \in {V} \}$

\While{$True$}
   \State $\ {P}_S \ \gets$ \parbox[t]{.75\linewidth}{%
take clusters from \(P\) in groups of size \(\min\bigl(n, \, |\text{remaining clusters in } P|\bigr)\) without replacement.

}
    \For{each ${P}_s \in  P_S $}
        \State ${V}' \gets$ union of all the clusters in ${P}_s$
        \State ${E}' \gets \text{get all edges relate to } V' \text{ in } E $
        \State $G'_s \gets ({V}', {E}')$
        \State $\text{SEs} \gets \emptyset$
        \State ${T}' \gets$ add a root tree node $\lambda$
        \For{each cluster ${C} \in {P}_s$}
            \State \parbox[t]{\dimexpr\linewidth-\algorithmicindent}{%
Add a node $\alpha$ to ${T}'$, s.t. $\alpha^{-} = \lambda$, $T_{\alpha} = {C}$}

            \For{each sentence $s \in {C}$}
                \State Add a leaf node $\gamma$ to ${T}'$, s.t.
                \\ \hskip6em $\gamma^{-} = \alpha$, $T_{\gamma} = \{s\}$
            \EndFor
        \State Caculate $SE_\alpha$ via Eq.~(3) 

        \State Append $SE_\alpha$ to SEs
        \EndFor
         \While{$True$}
            \State ${P}' \gets \{ \alpha \mid \alpha \in {T},\ h(\alpha) = 1 \}$
            \State $\Delta SE \gets \infty$
            \For{each $\alpha_i \in {P}$}
                \For{each $\alpha_j \in {P}$, $j > i$}
                    \State $\Delta SE_{ij} \gets$ Eq.~(4), w/o actually  \hspace*{13em} merging $\alpha_i$ and $\alpha_j$
                    \If{$\Delta SE_{ij} < \Delta SE$}
                        \State $\Delta SE \gets \Delta SE_{ij}$
                        \State $\alpha_{o_1} \gets \alpha_i$
                        \State $\alpha_{o_2} \gets \alpha_j$
                    \EndIf
                \EndFor
            \EndFor
            \If{$\Delta SE < 0$}
                \State \textproc{MERGE}($\alpha_{o_1}, \alpha_{o_2}$)
            \Else
                \State \textbf{Break}
            \EndIf
        \EndWhile
        \State Append ${P}'$ to ${P}$

    \EndFor
    \If{$ |P_S|  = 1$}
        \State \textbf{break}
    \EndIf
    \If{$P$ is unchanged since the last iteration}
        \State $n \gets 2n$
    \EndIf
\EndWhile

\State \Return ${P}$
\end{algorithmic}
\end{algorithm}

Then, we apply a clustering algorithm based on 2D SE minimization to the integrated graph $G'=\{V, E\}$, as shown in \textbf{Algorithm 1}. Specifically, each sentence initially resides in its own cluster (line 1). Subsequently, these clusters are combined into subsets of size $n$ (line 3), and each subset is transformed into a subgraph ${G}'_s = \{ {V}',  {E}' \}$ (lines 5-7). Then, each subgraph is converted into an encoding tree ${T}'$ (lines 9-14). The constructed encoding tree adheres to the following definition \cite{li2016structural}:
\begin{enumerate}
    \item Let ${T}'$ be a encoding tree structure over the set ${V}'$, where each node $\alpha$ in ${T}'$ is linked to a subset ${T}'_{\alpha} \subseteq {V}'$. The root node is denoted by $\lambda$, contains the entire set, i.e., ${T}'_{\lambda} = {V}'$. In contrast, each leaf node $\gamma$ corresponds uniquely to a single node from ${V}'$, such that ${T}'_{\gamma} = \{v\}$ for some $v \in {V}'$.
    
    \item For any node $\alpha$ in ${T}'$ with children $\beta_1, \ldots, \beta_k$, the sets ${T}'_{\beta_1}, \ldots, {T}'_{\beta_k}$ form a partition of ${T}'_{\alpha}$.
    
    \item For every node $\alpha$ in ${T}'$, we define its height as $h(\alpha)$. The height of a leaf node $\gamma$ is set to zero, i.e., $h(\gamma) = 0$. For any node $\alpha$, the height of its parent $\alpha^{-}$ is defined as one plus the height of $\alpha$, that is, $h(\alpha^{-}) = h(\alpha) + 1$. The height of ${T}'$, $h({T}') = h(\lambda)$.
\end{enumerate}

By applying Eq.~(2) and Eq.~(3), the $SE$ of each node $\alpha$ is calculated, after which the computed $SE_\alpha$ values are appended to the set $\text{SEs}$ (lines 15-16). 

\begin{equation}
{H}^{(1)}_\alpha = - \sum_{\substack{i=1 }}^{|{T}_\alpha|} \frac{d_i}{\mathit{vol}(\lambda)} \mathit{log} \frac{d_i}{\mathit{vol}(\lambda)}
\end{equation}
\begin{equation}
SE_\alpha = - \frac{g_{\alpha}}{\mathit{vol}(\lambda)} \mathit{log} \frac{\mathit{vol}(\alpha)}{\mathit{vol}(\lambda)} + {H}^{(1)}_\alpha
\end{equation}
where $\mathit{vol}(\lambda)$ denotes the sum of degrees of all nodes within $T_{\lambda}$, $g_\alpha$ represents the total weight of the cut edges by $T_\alpha$, and $d_i$ indicates the degree of node $i$, where $i \in T_\alpha$.

Under a greedy strategy, 2D SE minimization iteratively merges the pair of nodes in ${T}'$ that produces the largest $|\Delta SE|$, until no merge yields a negative $\Delta SE$ (lines 17–30), The calculation of  $\Delta SE$ between two nodes is as follows:
\begin{equation}
\begin{aligned}
\Delta{SE}_{i,j}=&{SE}_{new}-{SE}_{old} \\
=&{SE}_{\alpha_n}-({SE_{\alpha_{i}}} + {SE_{\alpha_{j}}})\\
=&-\frac{g_{\alpha_{n}}}{vol(\lambda)}log\frac{vol(\alpha_{n})}{vol(\lambda)}-\frac{vol(\alpha_{i})}{vol(\lambda)}log\frac{vol(\alpha_{i})}{vol(\alpha_{n})} \\
&- \frac{vol(\alpha_{j})}{vol(\lambda)}log\frac{vol(\alpha_{j})}{vol(\alpha_{n})}+\frac{g_{\alpha_{i}}}{vol(\lambda)}log\frac{vol(\alpha_{i})}{vol(\lambda)}& \\
&+\frac{g_{\alpha_{j}}}{vol(\lambda)}log\frac{vol(\alpha_{j})}{vol(\lambda)}.
\end{aligned}
\end{equation}
where $\alpha_{i}$, $\alpha_{j}$ is non-root nodes in encoding tree ${T}'$, the operation $MERGE(\alpha_i, \alpha_j)$ inserts a new node $\alpha_n$ into ${T}'$ and remove $\alpha_i$ and $\alpha_j$ from ${T}'$. $\alpha_n$ satisfies: 1) its children consist of all the children of $\alpha_i$ and $\alpha_j$; 2) its parent is the root node, i.e., $\alpha_n^{-} = \lambda$. The process is repeated until no more clusters can be merged. (lines 32-33). If no clusters can be merged in any subset, $n$ is increased so that each subset contains more clusters, making further merges possible (lines 34–35). Finally, a complete encoding tree ${T}$ of the integrated graph $G'$ is generated, and the leaf nodes of each subtree of the root node in $T$ form a sentence cluster. We denote the final clustering result as $\{C_1, C_2, .., C_c\}$.
\subsection{Summary Extraction Based on Position-Aware Compression}

To further extract summary sentences reflecting the core content of the input multi-document based on the above sentence clusters, we propose a position-aware compression method for summary extraction, as shown in Figure \ref{fig:Fig3}C.

Specifically, for the $j$-th sentence $s_{ij}$ in the $i$-th original document, its positional importance is defined as:
\[
\begin{gathered}
w_{pos}(s_{i,j})=1-\frac{pos_{i,j}-1}{N_i-1} \\
\end{gathered}
\]
where $pos_{ij}$ represents the position index of the sentence in original document, and $N_i$ denotes the total number of sentences in that document. Subsequently, we calculate the sum of $w_{pos}$ for all sentences in each sentence cluster (e.g., $C_i$) to obtain the positional-aware importance $Score_{pos}(C_i)$ of that cluster:
\[
\begin{gathered}
Score_{pos}(C_i) = \sum_{s \in Ci}w_{pos}(s)
\end{gathered}
\]
Then, we rank all clusters based on this metric to obtain the sorted clusters $\{C'_1, C'_2, ..., C'_{c}\}$, and retain the top $c^*$ clusters, namely $\{C'_1, C'_2, ..., C'_{c^*}\}$. Finally, following the approach of \citet{boudin2013keyphrase}, we use the K-shortest paths algorithm to find $K$ shortest paths as compression candidates $\{su^{(i)}_k\}_{k=1}^{K}$ in cluster $C'_i$. We calculate scores based on keyphrases to rank the compression candidates, and concatenate the Top-$1$ compression candidate from each cluster in sequence to obtain the final summary.

\section{Experimental Settings}\label{sec:setup}
\subsection{Datasets}
We conducted experiments on four datasets. DUC-2004 and Multi-News \cite{fabbri2019multi} are two commonly used news MDS datasets, primarily used to validate the effectiveness of MRGSEM-Sum. Additionally, to further validate the generalization ability of MRGSEM-Sum, we also conducted experiments on PubMed \cite{cohan2018discourse} and WikiSum \cite{liu2018generating}. PubMed is a dataset consisting of long documents, while WikiSum is a multi-document dataset from a non-news domain. Detailed statistics of these four datasets can be found in Table~\ref{tab:dataset}.
\begin{table}[htb!]
\fontsize{0.25cm}{0.4cm}\selectfont
    \caption{Test dataset statistics.}\label{tab:dataset}
\centering
\setlength{\tabcolsep}{1.2mm}
\begin{tabular}{@{}lcccc@{}}
\toprule
\multicolumn{1}{c}{\textbf{Datasets}} & \textbf{Test Dataset Size} & \textbf{\begin{tabular}[c]{@{}c@{}}Average Document\\ Length\end{tabular}} & \textbf{\begin{tabular}[c]{@{}c@{}}Docs Per\\ Cluster\end{tabular}} & \textbf{\begin{tabular}[c]{@{}c@{}}Average Summary\\ Length\end{tabular}} \\ \midrule
Multi-News                            & 5622                       & 2103.49                                                                    & 2.76                                                                & 263.66                                                                    \\
DUC-2004                              & 50                         & 5978.2                                                                     & 10.0                                                                & 107.04                                                                    \\
PubMed                                & 5025                       & 3224.5                                                                     & 1.0                                                                 & 209.5                                                                     \\
WikiSum                               & 2000                       & 2238.2                                                                     & 40.0                                                                & 101.2                                                                     \\ \bottomrule
\end{tabular}
\end{table}

\subsection{Automatic Evaluation Metrics}
In the model's automatic evaluation process, we followed the setup by Fabbri et al.\cite{fabbri2019multi} and used the automatic metric ROUGE to assess the quality of the final generated summaries. Specifically, we reported three key metrics: ROUGE-1 (R-1), ROUGE-2 (R-2), and ROUGE-SU (R-SU). ROUGE-1 and ROUGE-2 represent the n-gram overlap between the generated summary and the reference summary, while ROUGE-SU further integrates skip-gram and unigram considerations to evaluate the similarity between the generated summary and the reference summary.

\section{Experimental Results and Analysis}\label{sec:exper}

\subsection{E1. Comparative Experiments}

We compared MRGSEM-Sum with a series of strong baseline methods. Specifically, we selected two graph-based unsupervised extractive methods, \textbf{LexRank} \cite{erkan2004lexrank} and \textbf{TextRank} \cite{mihalcea2004textrank}, which capture key information in the document through graph models. In addition, we considered two similarity-based unsupervised methods, \textbf{MMR} \cite{carbonell1998use} and \textbf{Centroid} \cite{radev2004centroid}, which focus on computing the similarity between text units (such as sentences) to extract summaries. To further enrich the comparison scope, we introduced several supervised strong baseline models, including \textbf{PG-Original} \cite{see2017get}, \textbf{PG-MMR} \cite{lebanoff2018adapting}, \textbf{Copy-Transformer} \cite{gehrmann2018bottom}, and \textbf{Hi-MAP} \cite{fabbri2019multi}. Furthermore, given the advantages of current large language models (LLMs) in text generation, we specifically selected \textbf{GPT-3.5-turbo-16k} and \textbf{GPT-4-32k}, two LLMs capable of handling long texts, to test our method against LLMs.

The experimental results are shown in Table~\ref{tab:Comparison-1} . We can observe that: (1) MRGSEM-Sum outperforms other unsupervised models. Specifically, on Multi-News, compared to the classic graph-based unsupervised extractive baseline model LexRank, MRGSEM-Sum achieved improvements of over 4.94, 1.96, and 3.79 on R-1, R-2, and R-SU, respectively. There are also substantial improvements compared to the similarity-based unsupervised methods MMR and Centroid. Additionally, compared to the powerful graph clustering baseline model SummPip, MRGSEM-Sum also shows improvement, with increases of 0.89, 1.38, and 0.79 on R-1, R-2, and R-SU, respectively, and our model does not require pre-specifying the number of clusters in the extraction process. (2) Compared to supervised methods, MRGSEM-Sum also demonstrates strong competitiveness. Notably, on Multi-News, MRGSEM-Sum even outperforms the PG-Original and PG-MMR models, with an increase of 1.36 on R-1 compared to PG-Original, and an increase of 2.66 on R-1 compared to PG-MMR. These results indicate that our model achieves performance comparable to supervised models without the need for large-scale training data or complex training processes. (3) In the experiments with LLMs, we called the online API interfaces of these LLMs, and our standardized prompt was "Summarize the above article:". From the experimental results, it can be seen that the performance of MRGSEM-Sum  on the Multi-News and DUC-2004 datasets is comparable to that of LLMs, while our model is more compact.

\begin{table}[htb!]
    \caption{The comparison between MRGSEM-Sum and baselines on news MDS datasets Multi-News and DUC-2004 (\textbf{Bold} values are optimal among all methods; \underline{underlined} values are best among unsupervised methods).}
    \label{tab:Comparison-1}
    \resizebox{\linewidth}{!}{
\centering
    \setlength{\tabcolsep}{2.5mm}
    {\begin{tabular}{@{}lcccccc@{}}
\toprule
\multicolumn{1}{c}{\multirow{2}{*}{\textbf{Models}}} & \multicolumn{3}{c}{\textbf{Multi-News}}          & \multicolumn{3}{c}{\textbf{DUC-2004}}            \\ \cmidrule(l){2-7} 
\multicolumn{1}{c}{}                                 & \textbf{R-1}   & \textbf{R-2}   & \textbf{R-SU}  & \textbf{R-1}   & \textbf{R-2}   & \textbf{R-SU}  \\ \midrule
\multicolumn{7}{c}{\textbf{Unsupervised}}                                                                                                                  \\ \midrule
LexRank                                              & 38.27          & 12.70          & 13.20          & 35.56          & 7.87           & 11.86          \\
TextRank                                             & 38.44          & 13.10          & 13.50          & 33.16          & 6.13           & 10.16          \\
MMR                                                  & 38.77          & 11.98          & 12.91          & 30.14          & 4.55           & 8.16           \\
Centroid                                             & 41.25          & 12.83          & 15.63          & 36.52          & 8.82           & 11.68          \\
SummPip                                              & 42.32          & 13.28          & 16.20          & 36.30          & 8.47           & 11.55          \\ 
MRGSEM-Sum                                          & \underline{43.21} & \underline{14.66} & \underline{16.99} & \underline{\textbf{49.23}} & \underline{\textbf{15.36}} & \underline{\textbf{23.47}} \\
\midrule
\multicolumn{7}{c}{\textbf{Supervised}}                                                                                                                    \\ \midrule
PG-Original                                          & 41.85          & 12.91          & 16.46          & 31.43          & 6.03           & 10.01          \\
PG-MMR                                               & 40.55          & 12.36          & 15.87          & 36.42          & 9.36           & 13.23          \\
Copy-Transformer                                     & \textbf{43.57}          & 14.03          & 17.37          & 28.54          & 6.38           & 7.22           \\
Hi-MAP                                               & 43.47          & \textbf{14.89}          & \textbf{17.41}          & 35.78          & 8.90           & 11.43          \\ \midrule
\multicolumn{7}{c}{\textbf{LLMs}}                                                                                                                           \\ \midrule
GPT-3.5-turbo-16k                                    & 37.78          & 9.77           & 12.98          & 35.27          & 10.22          & 11.60          \\
GPT-4-32k                                            & 39.87          & 11.54          & 14.17          & 36.44          & 11.92          & 11.34          \\
\bottomrule
\end{tabular}}}
\end{table}

\subsection{E2. The Performance in Long Documents and Non-news Domain MDS Datasets.}
In order to further validate the generalization ability of MRGSEM-Sum, we conducted comparative experiments on PubMed and WikiSum datasets with unsupervised methods LexRank, TextRank, Centroid, and SummPip, as well as LLMs GPT-3.5-turbo-16k and GPT-4-32k.

The experimental results are shown in Table \ref{tab:Comparison-2}. We can observe that: (1) MRGSEM-Sum consistently outperforms unsupervised baseline models on these two datasets, particularly demonstrating significant improvements on the PubMed dataset. Compared to the Centroid, MRGSEM-Sum achieved improvements of 11.34, 8.47, and 8.1 on R-1, R-2, and R-SU, respectively. Even when compared with the powerful unsupervised baseline model SummPip, MRGSEM-Sum still achieved substantial performance gains. These results indicate that MRGSEM-Sum not only exhibits significant advantages on traditional news multi-document summarization datasets but also demonstrates strong adaptability and generalization ability on long document and non-news domain multi-document datasets like PubMed and WikiSum. (2) The performance of MRGSEM-Sum is comparable to LLMs and, in some cases, even superior. For instance, on the WikiSum dataset, our model showed an improvement of 2.52 on the R-1 compared to GPT-4-32k. This further confirms the advantages of MRGSEM-Sum over LLMs.

\begin{table}[htb!]
    \caption{The comparison between MRGSEM-Sum and baselines on PubMed and WikiSum (\textbf{Bold} values are optimal among all methods; \underline{underlined} values are best among unsupervised methods).}
    \label{tab:Comparison-2}
    \resizebox{\linewidth}{!}{
\centering
    \setlength{\tabcolsep}{2.5mm}
    {\begin{tabular}{@{}lcccccc@{}}
\toprule
\multicolumn{1}{c}{\multirow{2}{*}{\textbf{Models}}} & \multicolumn{3}{c}{\textbf{PubMed}}              & \multicolumn{3}{c}{\textbf{WikiSum}}             \\ \cmidrule(l){2-7} 
\multicolumn{1}{c}{}                                 & \textbf{R-1}   & \textbf{R-2}   & \textbf{R-SU}  & \textbf{R-1}   & \textbf{R-2}   & \textbf{R-SU}  \\ \midrule
\multicolumn{7}{c}{\textbf{Unsupervised}}                                                                                                                  \\ \midrule
LexRank                                              & 28.77          & 8.16           & 7.44           & 36.60          & 9.40           & 13.03          \\
TextRank                                             & 27.73          & 7.47           & 6.88           & 36.89          & 8.79           & 12.98          \\
MMR                                                  & 28.31          & 8.54           & 7.40           & 32.91          & 6.85           & 9.98           \\
Centroid                                             & 26.62          & 6.57           & 6.32           & 36.37          & 8.08           & 12.53          \\
SummPip                                              & 29.67          & 12.63          & 7.46           & 37.84          & 9.95           & 13.11          \\ 
MRGSEM-Sum                                           & \underline{37.96} & \underline{\textbf{15.04}} & \underline{14.32} & \underline{\textbf{39.07}} & \underline{\textbf{10.91}} & \underline{\textbf{14.06}} \\
\midrule
\multicolumn{7}{c}{\textbf{LLMs}}                                                                                                                           \\ \midrule
GPT-3.5-turbo-16k                                    & 39.30    & 11.60          & \textbf{14.57}          & 32.15          & 6.46           & 9.69           \\
GPT-4-32k                                            & \textbf{41.78}    & 13.35          & 13.35          & 36.55          & 9.06           & 12.29          \\
 \bottomrule
\end{tabular}}}
\end{table}

\subsection{E3. Ablation Experiments}
To better understand the contributions of different modules to the model performance, we conducted ablation experiments on Multi-News and PubMed datasets. Firstly, we replaced the multi-relation graph with the approximate discourse graph (ADG) used in SummPip (i.e., w/o multi-relation graph). Secondly, we replaced the 2D SE with spectral clustering method and set the number of clusters according to the experimental configuration by \citet{zhao2020summpip} (i.e., w/o 2D SE clustering). Lastly, we replaced the position-aware compression method with a more general compression method proposed by \citet{boudin2013keyphrase} (i.e., w/o position-aware).

The results are shown in Table~\ref{tab:Ablation}, it can be observed that when removing the multi-relation graph, 2D SE clustering, and position-aware modules, MRGSEM-Sum's R-1, R-2, and R-SU scores all exhibited varying degrees of decline. Removing the multi-relation graph and 2D SE clustering modules led to significant performance drops on both datasets. For instance, on the Multi-News dataset, removing the multi-relation graph module resulted in a 4.25 decrease in R-1, while removing the 2D SE clustering module led to a 5.35 decrease in R-1. Additionally, we noticed that removing the position-aware module had a more significant impact on the PubMed dataset, which we speculate is due to PubMed being a dataset of long documents, making it more sensitive to positional information during the summarization process.

\begin{table}[htb!]
    \caption{Ablation experiments on Multi-News and PubMed.}
    \label{tab:Ablation}
    \resizebox{\linewidth}{!}{
\centering
    \setlength{\tabcolsep}{1.8mm}
    {\begin{tabular}{@{}lcccccc@{}}
\toprule
\multicolumn{1}{c}{\multirow{2}{*}{\textbf{Models}}} & \multicolumn{3}{c}{\textbf{Multi-News}}          & \multicolumn{3}{c}{\textbf{PubMed}}              \\ \cmidrule(l){2-7} 
\multicolumn{1}{c}{}                                 & \textbf{R-1}   & \textbf{R-2}   & \textbf{R-SU}  & \textbf{R-1}   & \textbf{R-2}   & \textbf{R-SU}  \\ \midrule
MRGSEM-Sum                                           & \textbf{43.21} & \textbf{14.66} & \textbf{16.99} & \textbf{37.96} & \textbf{15.04} & \textbf{14.32} \\
w/o multi-relation graph                             & 38.96          & 11.86          & 13.54          & 30.77          & 11.35          & 9.60           \\
w/o 2D SE clustering                                 & 37.86          & 11.65          & 13.05          & 34.94          & 14.12          & 11.45          \\
w/o position-aware                                   & 42.06          & 14.12          & 15.69          & 23.37          & 11.97          & 4.20           \\ \bottomrule
\end{tabular}}}
\end{table}

\subsection{E4. The Impact of Position-aware Compression}

In ablation experiment, we have validated the effectiveness of position-aware compression. Next, we designed an experiment to further explore the impact of position-aware compression. Specifically, for the clusters $\{C'_1, C'_2, ..., C'_{c^*}\}$ sorted and truncated based on position-aware importance, we selected $c^* = 7, 8,...,19$ as inputs to the method proposed by \citet{boudin2013keyphrase}, and observed the quality of the generated summaries. The results are presented in Table~\ref{tab:Noise Ratios}, where it can be observed that under the position-aware importance sorting, even when only a portion of clusters participated in the subsequent summary generation, it still yielded satisfactory results. Additionally, an increase in the number of clusters led to a decrease in summary quality. This once again underscores the necessity of position-aware compression.
\begin{table}[htb!]
\caption{The Impact of Position-aware Compression}
\centering
\resizebox{0.5\columnwidth}{!}{%
\begin{tabular}{cccc}
\toprule
                        & \multicolumn{3}{c}{\textbf{Multi-News}}                                 \\ \cline{2-4} 
\multirow{-2}{*}{\textbf{$c^*$}} & \textbf{R-1}            & \textbf{R-2}                          & \textbf{R-SU}            \\ \hline
ALL                     & 42.06          & 14.12 & 15.69          \\ \hline
7                       & 39.66          & 12.14                        & 14.31          \\
8                       & 41.01          & 12.77                        & 15.35          \\
9                       & 41.93          & 13.25                        & 16.06          \\
10                      & 42.60          & 13.69                        & 16.58          \\
11                      & 42.97          & 13.99                        & 16.85          \\
12                      & 43.18          & 14.26                        & 16.98          \\
13             & \textbf{43.26} & 14.50                        & 16.90          \\
14                      & 43.21          & 14.66                        & \textbf{16.99} \\
15                      & 43.09          & 14.78                        & 16.75          \\
16                      & 42.93          & 14.90                        & 16.58          \\
17                      & 42.74          & 14.97                        & 16.38          \\
18                      & 42.57          & 15.03                        & 16.20          \\
19                      & 42.43          & \textbf{15.06}               & 16.05          \\ \bottomrule
\end{tabular}%
}

\label{tab:Noise Ratios}
\end{table}

\subsection{E5. Human Evaluation}
In addition to automatic evaluations, we also conducted human evaluations for MRGSEM-Sum. We recruited three graduate students with excellent English reading and writing skills as evaluators, who independently rated 100 randomly selected summaries generated by each model in the Multi-News test set, including TextRank, Centroid, SummPip, GPT-3.5-turbo-16k, GPT-4-32k, and MRGSEM-Sum. The evaluators were instructed to rate the generated summaries based on three criteria: Fluency, Consistency, and Coverage, using a 5-star scale where 1 star represents the worst performance and 5 stars the best, and the average score from the evaluators represents the final score.

\begin{figure}
  \centering
  \includegraphics[width=0.9\linewidth]{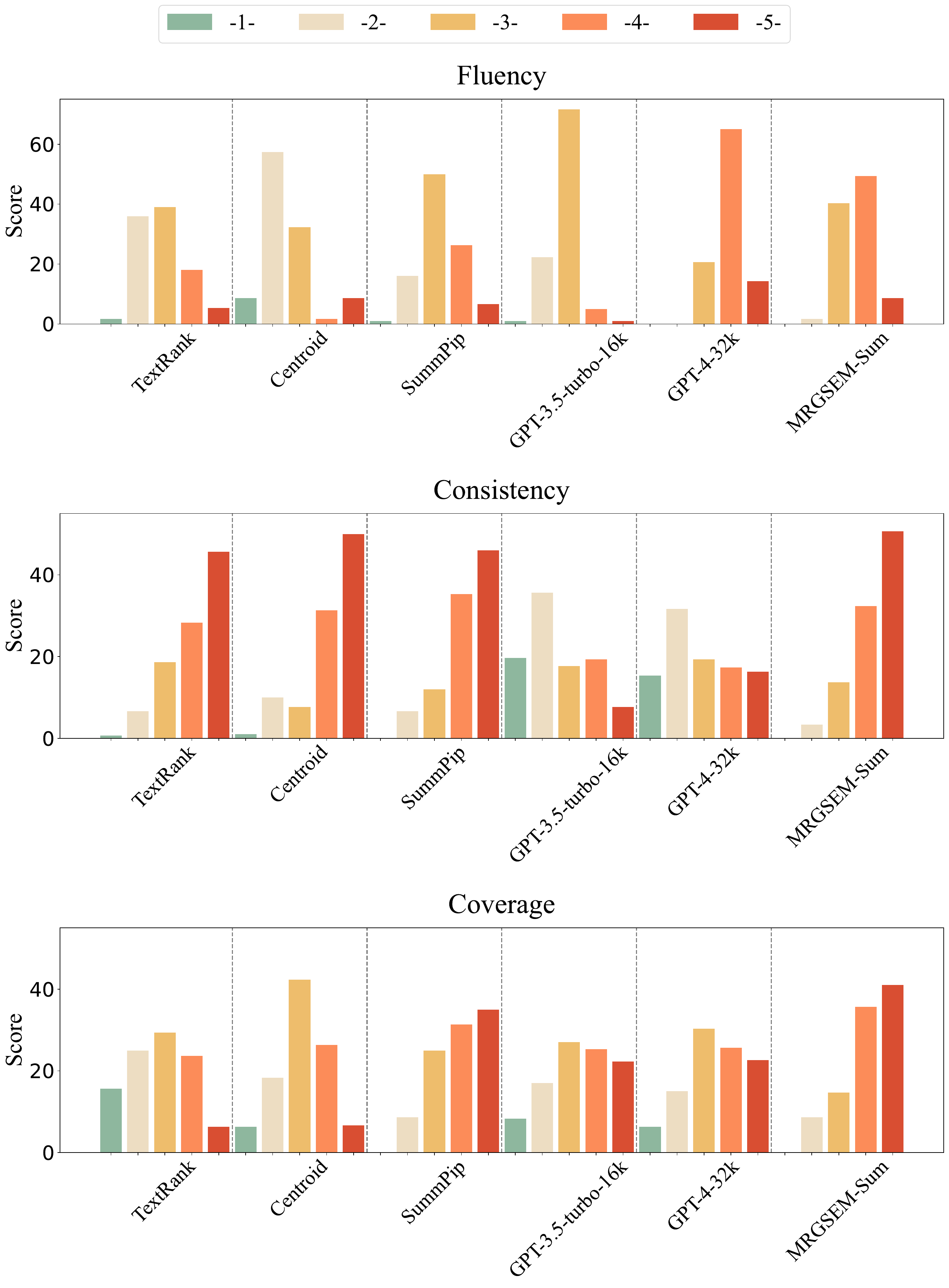}
  \caption{Human Evaluation.}
  \label{fig:myfig4}
\end{figure}

Figure~\ref{fig:myfig4} displays the distribution of human evaluation scores for these six models. It can be observed that MRGSEM-Sum has a significantly higher proportion of 5-star ratings in Consistency and Coverage compared to the other methods, particularly LLMs. In terms of Fluency, the performance of MRGSEM-Sum is comparable to LLMs and significantly better than other unsupervised summarization models. These findings further validate the effectiveness of the proposed MRGSEM-Sum. Additionally, an interesting observation is that both the cluster-based unsupervised summarization model SummPip and MRGSEM-Sum exhibit excellent performance in the Coverage metric, indicating that the approach of clustering sentences first and then compressing them can significantly enhance the coverage of generated summaries.

\subsection{E6. Case Study}
To provide a more intuitive demonstration of the performance of MRGSEM-Sum, we conducted a case study. We randomly selected an example from the results generated by the baseline model and MRGSEM-Sum on the Multi-News test set for qualitative analysis. 

\begin{figure}[h!]
  \centering
  \includegraphics[width=\linewidth]{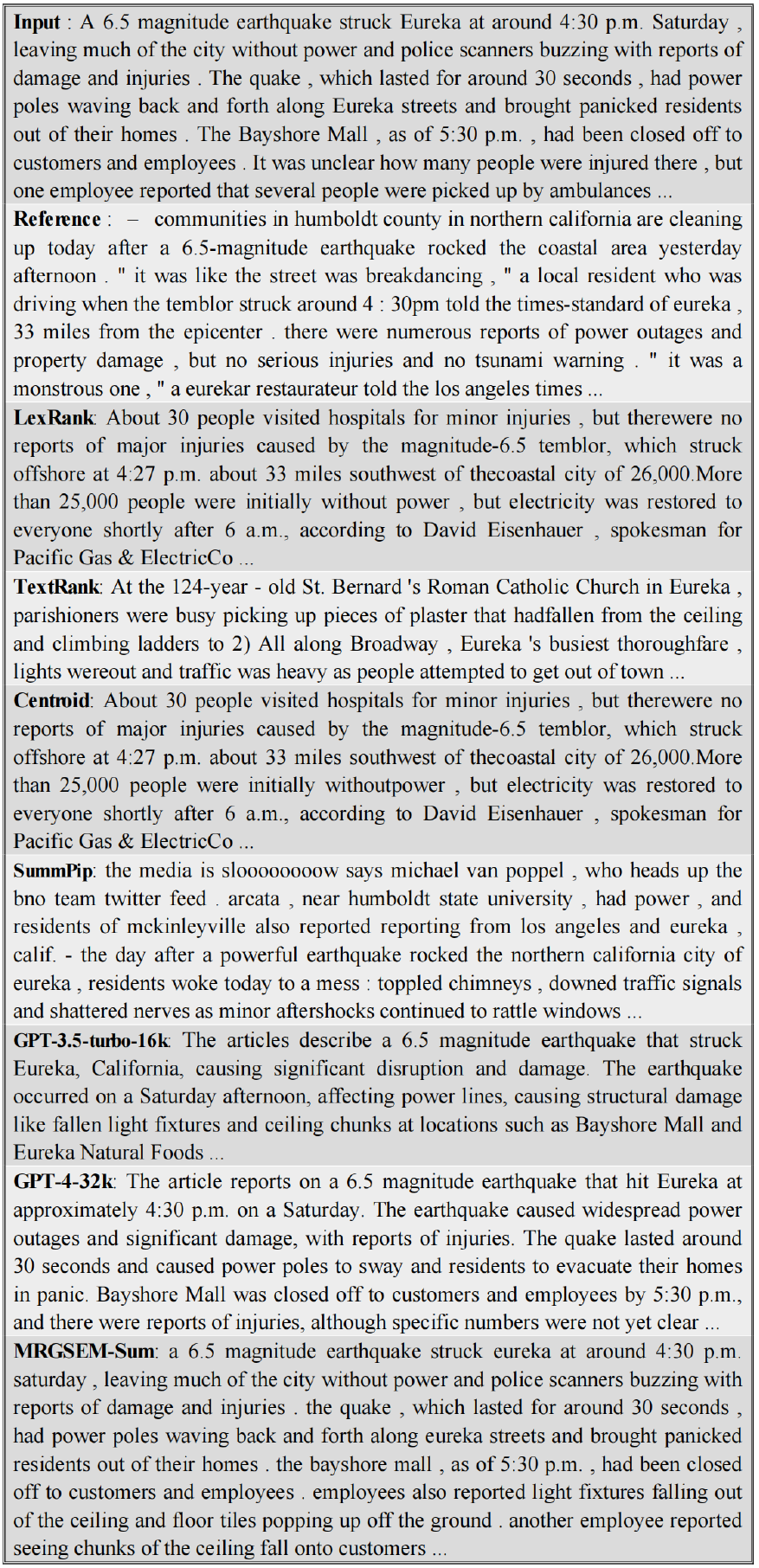}
  \caption{Case Study.}
  \label{fig:myfig5}
\end{figure}

As shown in Figure \ref{fig:myfig5}, compared to other models, the summary generated by MRGSEM-Sum contains more detailed information and is closer to the reference summary. For example, the summary generated by MRGSEM-Sum includes key event elements, such as ``A 6.5-magnitude earthquake struck eureka at around 4:30 p.m. saturday'' and ``without power''. These elements are highly consistent with both the input documents and the reference summary. Furthermore, MRGSEM-Sum also captures additional important details, such as ``around 30 seconds'' and ``as of 5:30 p.m.'', which, although not in the reference summary, are present in the input documents and relatively significant. Even compared to LLs, the summary generated by MRGSEM-Sum is more informative. In contrast, unsupervised summarization models like SummPip miss key elements, such as ``A 6.5-magnitude earthquake'', and extract irrelevant information not related to the news topic, such as ``the media is sloooooooow''. 

\section{Conclusion}\label{sec:conclu}
We present MRGSEM-Sum, an unsupervised multi-document summarization framework that addresses the dual challenges of complex relationships and information redundancy in MDS through three key innovations: (1) a multi-relational graph integrating both semantic and discourse relations to model sentence connections more comprehensively than single-relation approaches; (2) a 2D structural entropy minimization algorithm for adaptive, parameter-free clustering that automatically groups sentences while minimizing redundancy; and (3) a position-aware compression mechanism to distill salient information from clusters. Extensive experiments on Multi-News, DUC-2004, PubMed, and WikiSum demonstrate that MRGSEM-Sum outperforms unsupervised baselines and rivals supervised methods and LLMs in summary quality, as validated by ROUGE scores and human evaluations.

\section{Acknowledgments}
Thank the reviewers for their meticulous review of this paper.

\bibliography{aaai2026}

\end{document}